\documentclass[final]{cvpr}

\usepackage{times}
\usepackage{epsfig}
\usepackage{graphicx}
\usepackage{subfigure}
\usepackage{amsmath}
\usepackage{amssymb}
\usepackage{multirow}
\usepackage{booktabs}
\usepackage{bbding}
\usepackage{enumitem}
\usepackage{footmisc}
\usepackage{overpic}
\usepackage{bm}

\usepackage{color}

\usepackage[pagebackref=true,breaklinks=true,colorlinks,bookmarks=false]{hyperref}

\def\cvprPaperID{7889} % *** Enter the CVPR Paper ID here
\def\confYear{CVPR 2021}
\begin{document}

%%%%%%%%% TITLE
\title{Towards Diverse Paragraph Captioning for Untrimmed Videos}

\author{Yuqing Song\textsuperscript{1}\footnotemark[1], Shizhe Chen\textsuperscript{2}\footnotemark[1], Qin Jin\textsuperscript{1}\footnotemark[2]\\
\textsuperscript{1}Renmin University of China, \textsuperscript{2}INRIA\\
{\tt\small syuqing@ruc.edu.cn, cshizhe@gmail.com, qjin@ruc.edu.cn}
% \and
}

\maketitle
\thispagestyle{empty}

\footnotetext[1]{Equal contribution. This work was performed when Shizhe Chen was at Renmin University of China.}
\footnotetext[2]{Corresponding author.}

%%%%%%%%% ABSTRACT
\begin{abstract}
Video paragraph captioning aims to describe multiple events in untrimmed videos with descriptive paragraphs.
Existing approaches mainly solve the problem in two steps: event detection and then event captioning.
Such two-step manner makes the quality of generated paragraphs highly dependent on the accuracy of event proposal detection which is already a challenging task.
In this paper, we propose a paragraph captioning model which eschews the problematic event detection stage and directly generates paragraphs for untrimmed videos.
To describe coherent and diverse events, we propose to enhance the conventional temporal attention with dynamic video memories, which progressively exposes new video features and suppresses over-accessed video contents to control visual focuses of the model.
In addition, a diversity-driven training strategy is proposed to improve diversity of paragraph on the language perspective.
Considering that untrimmed videos generally contain massive but redundant frames, we further augment the video encoder with keyframe awareness to improve efficiency.
Experimental results on the ActivityNet and Charades datasets show that our proposed model significantly outperforms the state-of-the-art performance on both accuracy and diversity metrics without using any event boundary annotations.
Code will be released at \url{https://github.com/syuqings/video-paragraph}.
\end{abstract}

%%%%%%%%% BODY TEXT
\section{Introduction}
Describing videos with natural language sentences, a.k.a. video captioning, has attracted increasing research attentions due to the rapid emergence of videos in our lives.
The dominant video captioning task \cite{pan2020spatio-temporal,zhang2020object-relation,venugopalan2015vtt,yao2015temporal,zanfir2016st} focuses on generating a single sentence to describe a carefully trimmed video which mainly contains one major event within short duration such as 10-20 seconds \cite{xu2016msrvtt,wang2019vatex}.
However, the videos in the wild are mostly untrimmed with rich temporal event structures.
A single sentence is insufficient to convey fine-grained information in such untrimmed videos.
Therefore, recent works \cite{xiong2018MFT,park2019AdvInf,lei2020MART} have attempted to generate a story-oriented paragraph with multiple sentences to comprehensively describe video contents.

% drawbacks of previous works: require event proposals
Existing works \cite{xiong2018MFT,park2019AdvInf,lei2020MART} mainly adopt a two-stage framework for video paragraph captioning: firstly detecting event segments in the video, and then generating the event description for each segment.
Despite being reasonable, the framework requires temporal segment coordinates for descriptions in the paragraph to train the model, which are expensive to annotate.
Moreover, since event categories are extremely diverse in open-domain untrimmed videos, it is quite challenging to detect precise event segments compared with the action detection task \cite{zhou2016multistage,zhao2017ssn,chao2018rethinking}, which has a fixed category list.
The poorly detected events greatly harm the performance of paragraph captioning in existing frameworks.
As a result, several works \cite{park2019AdvInf,lei2020MART} use ground-truth event segments to generate video paragraphs, which cannot generalize to videos without such event annotations.

However, \emph{is event detection really necessary for video paragraph captioning?}
Let's review a simpler task of image paragraph captioning. The state-of-the-art approaches \cite{luo2019paragraph,Melas2018paragraph} directly generate sentences from images without predicting sequences of image coordinates. The generated paragraphs have shown good capability to capture descriptive logic such as from foreground to background.
Motivated by these works, we aim to eschew the costly event segment detection process, and efficiently generate video paragraph descriptions in a single stage.

Compared with the image counterpart, there are mainly three challenges for video paragraph captioning when event segments are unavailable.
% efficiency -> keyframe selection
Firstly, an untrimmed video generally consists of hundreds or thousands of frames, while an image contains much fewer region candidates to be attended to. Therefore, it consumes more computation resources during description generation.
% coherent and event diversity, hard to learn attention -> controllable attention 
Secondly, the large number of frame candidates also makes it hard for the captioning model to learn an effective attention mechanism to form a coherent descriptive logic and describe diverse events in the video, especially when the training examples are limited.
% language diversity -> diversify loss
Thirdly, the captioning model usually tends to generate redundant words and phrases that are of high frequency in the dataset especially for the long paragraph generation.

% our model
In this work, we propose an one-stage framework to tackle the above challenges for diverse and efficient video paragraph generation.
Considering that there are many redundant frames in untrimmed videos, we propose to automatically select keyframes during the video encoding via additional video semantic summary loss and sparsity loss. In this way, only keyframes are used to generate the long paragraph during inference, which improves the computational efficiency.
To guide the model in effective description logic learning for diverse and coherent events, we propose to improve conventional temporal attention with dynamic video memory which tracks and controls visual focuses in the video. It includes an ``add'' operation to progressively expose new video frames to the model, and an ``erase'' operation to suppress over-accessed video contents.
To further improve diversity of generated paragraphs from language perspective, we improve the training objective with token-level and sequence-level high-frequency penalties to encourage generating more unique expressions.
Experiments show that our model outperforms two-stage methods which even utilize ground-truth event segments on ActivityNet dataset, and also achieves the state-of-the-art result on Charades dataset which does not have temporal annotations.

The main contributions of this work are as follows:
\parskip=0.1em
\begin{itemize}[itemsep=0.5pt,partopsep=0pt,parsep=\parskip,topsep=0.5pt]
    \item To the best of our knowledge, we are the first to eschew event detection stage and directly generate paragraphs for untrimmed videos, which avoids the dependence on expensive event temporal annotations.
    \item We propose an attention mechanism with dynamic video memories and diversity-driven training objectives to generate coherent and diverse paragraph from video and language perspectives, and improve generation efficiency via keyframe-aware video encoder.
    \item Our model achieves state-of-the-art results on both ActivityNet and Charades datasets without using any event boundary annotations.
\end{itemize}
\section{Background: Vanilla Paragraph Captioning}
\label{sec:background}
% task
Given an untrimmed video $v$, the video paragraph captioning task aims to generate a paragraph $y=\{y_1,\cdots,y_T\}$ to describe events in $v$, where $y_t$ denotes the $t$-th word in the paragraph.
In the following, we introduce a vanilla video paragraph captioning model without the event segment detection stage and discuss its limitations. 

% encoder and decoder
The vanilla model is similar to conventional video captioning models \cite{pan2020spatio-temporal,zhang2020object-relation,venugopalan2015vtt} based on the encoder-decoder framework \cite{sutskever2014encdec}.
The encoder transforms $v$ into a sequence of clip-level features. 
Specifically, we first divide $v$ into non-overlapping clips with 64 frames per clip and use pretrained CNNs \cite{he2016resnet,carreira2017i3d} to extract features for each clip as $\mathcal{X}^0=\{x^0_1,\cdots,x^0_L \}$, where $L$ is the number of clips.
To encode long-range temporal dependencies among clips, we apply $N$ transformer layers on $\mathcal{X}^0$ as follows:
\begin{equation}
   \mathcal{X}^i = \mathrm{FFN}(\mathcal{X}^{i-1} + \mathrm{MultiHead}(\mathcal{X}^{i-1},\mathcal{X}^{i-1},\mathcal{X}^{i-1}))
\end{equation}
where $\mathrm{FFN}(\cdot)$ and $\mathrm{MultiHead}(\cdot)$ denote feed-forward network and multi-head attention as in \cite{vaswani2017transformer}.
The hidden state $\mathcal{X}^N$ is used as the encoded video feature $\mathcal{V}^{enc} \in \mathbb{R}^{L\times d}$, where $d$ is the feature dimension.

For the decoder, we use $N$ layers of transformer due to the advantage of its structure in long text generation \cite{zhou2018endtoend,lei2020MART}.
Besides the self-attention as in the encoder, the decoder further adopts the cross-modal multi-head attention to compute attention weights on $\mathcal{V}^{enc}$ at each decoding step.
Therefore, each word is generated conditioning on previously predicted words and the attended video contents.

% training
The captioning model is typically trained by maximum likelihood estimation (MLE) given the ground-truth pair $(v, y^{*})$, where $y^{*}=\{y^*_1,\cdots,y^*_T\}$, which is:
\begin{equation}
\label{eqn:xe}
    \mathcal{L}_{\text{mle}} = - \frac{1}{T}\sum_{t=1}^T \log p(y^*_t|y^*_{<t},v)
\end{equation}
To address the exposure bias and target mismatch \cite{rennie:sc} problems in MLE, reinforcement learning (RL) \cite{ranzato2016rl} is usually adopted to further improve the model with sequence-level non-differentiable caption rewards as follows:
\begin{equation}
\label{eqn:rl}
    \mathcal{L}_{\text{rl}} = - \frac{1}{T}r(y^s)\sum_{t=1}^T \log p(y^s_t|y^s_{<t},v)
\end{equation}
where $y^s=\{y^s_1,\cdots,y^s_T\}$ is a paragraph sampled from the model and $r(\cdot)$ is the reward function.

Without event segment annotations, the above vanilla model suffers from three limitations for video paragraph captioning.
Firstly, the untrimmed video usually contains a large number of clips, while the vanilla encoder feeds all clip features to the following decoder which brings huge attention computation burden for long paragraph generation.
Secondly, due to the large amount of clip candidates and limited training examples, it is hard for the decoder to learn effective attention mechanism to form coherent descriptive logic.
Finally, both MLE and RL training make the model more likely to generate high-frequency words and phrases, and thus harm the diversity of generated paragraphs.
Therefore, it is essential to address these limitations to make the one-stage model more practical.
\section{The Proposed Method}
\begin{figure*}
  \centering
  \begin{overpic}[scale=0.43]{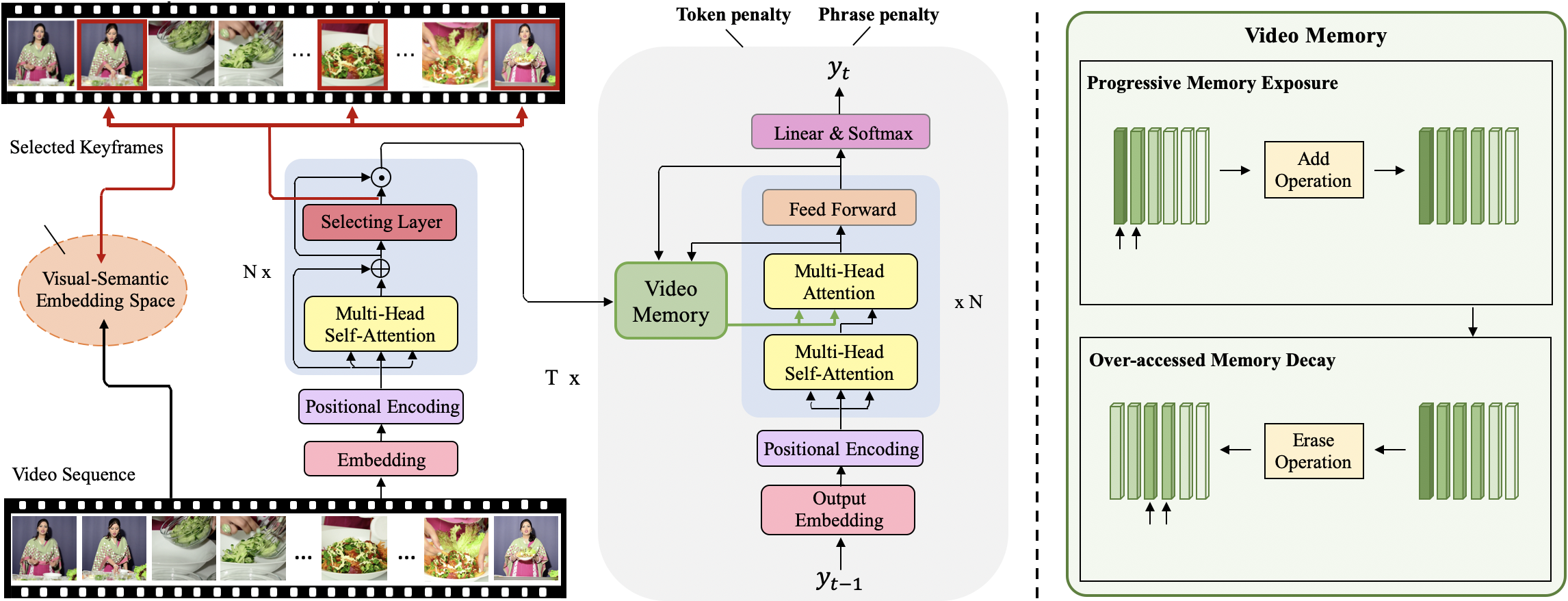}
    \put(7.4,24.2){\footnotesize{$\mathcal{X}^0_{key}$}}
    \put(7.4,14.6){\footnotesize{$\mathcal{X}^0$}}
    \put(0.3,25){\footnotesize{$\mathcal{L}_{\text{reconst}}$}}
    \put(34.3,20){\footnotesize{$\mathcal{V}^{enc}$}}
    \put(41.8,15){\footnotesize{$\mathcal{M}_t$}}
    \put(45.2,28.4){\footnotesize{$h_t$}}
    \put(45.2,23.5){\footnotesize{$\bm{\alpha}_t$}}
    \put(14.8,27.3){\footnotesize{$\bm{s}^N$}}
    \put(77.6,30.4){\footnotesize{$\mathcal{M}_t$}}
    \put(85.9,30.3){\footnotesize{$\mathcal{\hat M}_{t+1}$}}
    \put(85.9,12.5){\footnotesize{$\mathcal{\hat M}_{t+1}$}}
    \put(77.6,12.6){\footnotesize{$\mathcal{M}_{t+1}$}}
    \put(70.5,20.6){\footnotesize{$\hat m_{t+1,i} = m_{t,i} + g^a_t(1-u_{t,i})p^a_{t,i}v_i^{enc}$}}
    \put(72,3){\footnotesize{$m_{t+1,i} = \hat m_{t+1,i}(1 - g^e_t\tilde{\alpha}_{t,i}p^e_{t,i})$}}
  \end{overpic}
  \caption{\emph{Left}: The framework of our proposed video paragraph captioning model. \emph{Right}: Details of the proposed dynamic video memories with two updating mechanisms for description coherence and diversity respectively. $\oplus$ denotes addition and $\odot$ denotes hadamard product.}
  \label{fig:model}
  \vspace{-13pt}
\end{figure*}
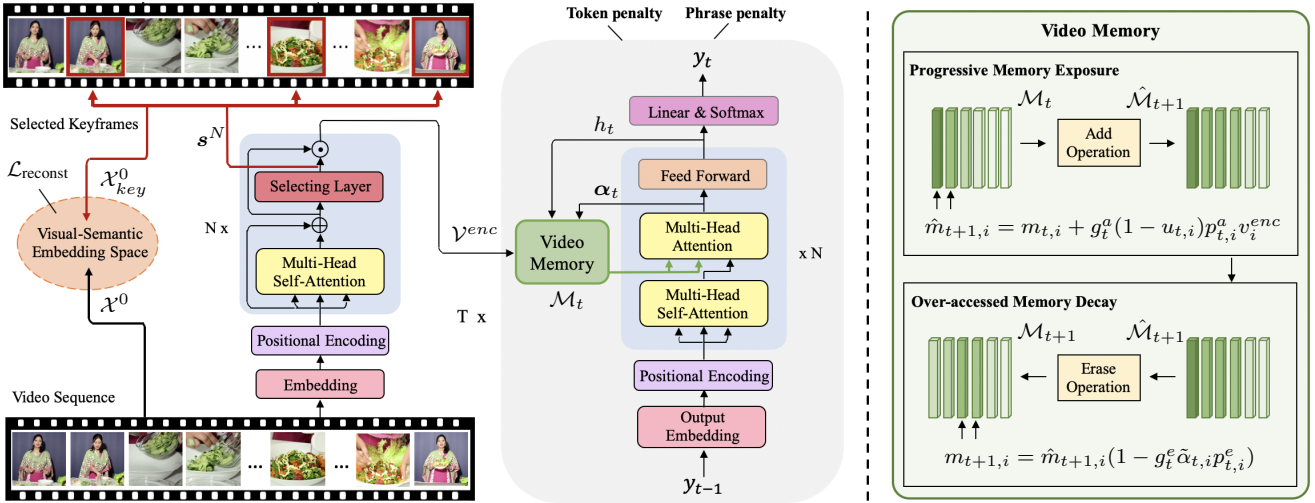

In this section, we introduce the proposed video paragraph captioning model, which is illustrated in Figure~\ref{fig:model}.
We first describe the keyframe-aware video encoder in Section~\ref{sec:keyframe}, which selects key frames during video encoding to improve attention computation efficiency in decoding.
Then in Section~\ref{sec:memory_update}, we present a more effective cross-modal attention mechanism enhanced by dynamic video memories, including \emph{progressive memory exposure} to guide the model describing temporally coherent events, and \emph{over-accessed memory decay} to reduce repetitions on described events.
Finally, we present the proposed diversity-driven training objective in Section~\ref{sec:objective}, which improves the language diversity via penalizing tokens and phrases of high-frequency.

\subsection{Keyframe-aware Video Encoder}
\label{sec:keyframe}
Considering that there is a large amount of clips in untrimmed videos which brings huge attention computation burden to the decoder, we propose to explicitly select key frames during video encoding.
We augment the vanilla encoder with a keyframe selection layer, which predicts the informativeness of each clip based on its contextual representation.
In the $i$-th encoding layer, the encoded video feature is computed as follows:
\begin{eqnarray}
    && \mathcal{\hat{X}}^i = \mathcal{X}^{i-1} + \mathrm{MultiHead}(\mathcal{X}^{i-1},\mathcal{X}^{i-1},\mathcal{X}^{i-1}) \\
    && \bm{s}^i = \sigma(\mathrm{FFN}(\mathcal{\hat{X}}^i)) \\
    && \mathcal{X}^i = \mathcal{\hat{X}}^{i} \cdot \bm{s}^i
\end{eqnarray}
where $\sigma$ is the sigmoid function, $s^i_j$ is a scalar to infer the informativeness of the $j$-th clip to the $i$-th encoding layer.
In this way, we can employ $\bm{s}^N \in \mathbb{R}^L$ to identify the key frames to be used in the following decoder.

However, the paragraph generation loss alone cannot provide sufficient supervision for the keyframe selection and the poorly selected video features can hinder effective attention learning in the decoder.
Since key frames can well represent the semantic contents of a whole video, they are expected to reconstruct original video in the semantic space.
Therefore, we propose to reconstruct high-level semantic embedding of the video in a visual-semantic joint embedding space.
Specifically, we utilize the video-text retrieval \cite{faghri2018vse++} task as a proxy task to pre-train such visual-semantic joint embedding space.
We first feed the video feature sequence $\mathcal{X}^0$ and the ground-truth paragraph $y^*$ to GRUs \cite{cho2014gru} respectively to get a global encoding vector for each of them.
Then they are mapped to the joint embedding space with the hard negative triplet loss \cite{faghri2018vse++} to ensure the video/text with similar semantics will be embeded closer.
After pre-training, we fix the video $\mathrm{GRU}^v$ and employ it to compute the video reconstruction loss as follows:
\begin{equation}
\label{eqn:reconstruct_loss}
    \mathcal{L}_{\text{reconst}} = \left\|\mathrm{GRU}^v( \mathcal{X}^0_{key})-\mathrm{GRU}^v(\mathcal{X}^0)\right\|_2
\end{equation}
where $\mathcal{X}^0_{key}=\mathcal{X}^0 \cdot \bm{s}^N$, which is the soft selected keyframe features.
We choose to reconstruct $\mathcal{X}^0$ rather than $y^*$ because there is less cross-modal gap between $\mathcal{X}^0_{key}$ and $\mathcal{X}^0$ making the learning more effective. 
To penalize a large number of key frames being selected, we further introduce a sparsity loss as follows:
\begin{equation}
\label{eqn:sparsity_loss}
    \mathcal{L}_{\text{sparsity}} = \left\|\frac{1}{L}\sum_{j=1}^Ls^N_j-\delta\right\|_1
\end{equation}
where $L$ is the total number of video clips and $\delta$ is the hyper-parameter which denotes the selecting ratio of key frames.
Notice that in the training phase, we use the soft selection of key frames for the gradient back-propagation, while at inference time, we use $\bm{s}^N$ to select the top $\lceil \delta L \rceil$ key frames to reduce the computational cost for decoding efficiency.

\subsection{Attention with Dynamic Video Memories}
\label{sec:memory_update}
With the encoded video feature sequence $\mathcal{V}^{enc}$, the decoder employs temporal attention mechanism to generate the paragraph.
However, the video paragraphs usually contain rich temporal logic structures, which are hard to learn by traditional attentions from limited training examples.
Therefore, we enhance the temporal attention mechanism in the decoder with dynamic video memories.

Instead of attending on the same $\mathcal{V}^{enc}$ at each decoding step, our model attends on video memory $\mathcal{M}_t=\{m_{t,1},\cdots,m_{t,L}\}$ at each step $t$, which are dynamically updated to make visual attentions moving coherently on diverse events.
Suppose $\bm{\alpha}_{t}\in\mathbb{R}^L$ (averaged from multiple heads and layers) is the overall attention weights over the video memory $\mathcal{M}_{t}$ at $t$-th step.
We utilize attention histories $\{\bm{\alpha}_{t-W}, \cdots, \bm{\alpha}_{t}\}$ with a history window $W$ to update $\mathcal{M}_t$ to $\mathcal{M}_{t+1}$ for the attention at next step.
We use the attention histories instead of $\bm{\alpha}_{t}$ because we expect to update the video memory when a complete phrase or sentence has been generated. The attention weights for a single word are noisy to indicate if the model should move to other frames.
We aggregate the attention histories into $\bm{\tilde{\alpha}}_{t} \in \mathbb{R}^L$ as follows to make attentions from more recent steps more important:
\begin{eqnarray}
    \label{eqn:W}
    & \bm{\tilde{\alpha}}_{t} = \sum_{j=0}^W w_{j}\bm{\alpha}_{t-j} \\
    & w_j = \frac{e^{1 -(j/W)}}{\sum_{k=0}^W e^{1 -(k/W)}}
\end{eqnarray}
where $w_j$ is the history decay weight.
The $\bm{\tilde{\alpha}}_{t}$ is then used to update the video memory via two operations, which include an ``add'' operation in \emph{progressive memory exposure} to progressively add more video clip features to the memory and an ``erase'' operation in \emph{over-access memory decay} to erase already described clips.

\textbf{Progressive Memory Exposure.}
To keep the event description coherent such as following the temporal order, we propose to progressively expose the video feature sequence $\mathcal{V}^{enc}$ to the attended video memory $\mathcal{M}$.
We first initialize the $\mathcal{M}_0$ as follows:
\begin{eqnarray}
    & \mathcal{M}_0 = \bm{u}_{0} \cdot \mathcal{V}^{enc} \\
    \label{eqn:S}
    & u_{0, i} = \begin{cases}1-(i/S),& i \leqslant S \\ 0,& i>S \end{cases}
\end{eqnarray}
where $\bm{u}_{t} \in \mathbb{R}^{L}$ denotes the exposure status at step $t$, which records the proportion of each clip feature added to the video memory. The $u_{t,i} \in [0, 1]$ is constantly updated, where $u_{t,i} = 1$ indicates that the $i$-th clip feature should not be added anymore.
$S$ is the initialization window length.
It can make the decoder focus on the beginning of video first, rather than randomly starting the paragraph generation.

We propose an adding gate $g^a_t \in [0,1]$ to determine whether we should ``add'' new features to the memory at step $t$.
Because when non-visual words are generated or the accessed video frame has not been fully described, the video memories should be updated less.
The gate $g^a_t$ is computed as follows:
\begin{equation}
    g^a_t = \sigma(f_{add}(h_t;\theta_{add}))
\end{equation}
where $\sigma$ is the sigmoid function, $h_t$ is the output hidden state at the $t$-th step and $f_{add}$ is a fully connected network parameterized by $\theta_{add}$.

Then we compute for each clip feature whether it needs to be added in the new video memory according to their visual relevance with previous context $\tilde{c}_t$.
The context can make the model describe events which are relevant to the previous one at next step to keep the event coherence.
The $\tilde{c}_t$ is computed based on aggregated attention history $\bm{\tilde{\alpha}}_{t}$ as follows:
\begin{equation}
    \tilde{c}_{t} = \sum_{i=1}^L\tilde{\alpha}_{t,i} \cdot v^{enc}_i
\end{equation}
Therefore, the probabilities of each clip feature to be added to the video memory is computed as:
\begin{equation}
    p^a_{t,i} = \sigma(f_{vis}([v^{enc}_i;\tilde{c_{t}}];\theta_{vis}))
\end{equation}
where $f_{vis}$ is the fully connected network similar to $f_{add}$.

Based on the adding gate $g^a_t$ and adding probability $p^a_{t,i}$ of each clip feature, we gradually add video features to the memories, which is:
\begin{eqnarray}
    & \hat m_{t+1,i} = m_{t,i} + g^a_t(1-u_{t,i})p^a_{t,i}v_i^{enc} \\
    & u_{t+1,i} = u_{t,i} + g^a_t(1-u_{t,i})p^a_{t,i}
\end{eqnarray}
where $\hat m_{t+1,i}$ is the intermediate memory feature which will be further processed with the ``erase'' operation.

\textbf{Over-accessed Memory Decay.}
In addition to the description coherence, describing diverse content of the video is also important to the video paragraph generation.
To prevent the decoder from only focusing on a few salient frames, we propose to weaken the already accessed features to encourage the model to describe more unseen video frames.
Similar to the progressive memory exposure mechanism, we employ $g^e_t$ as an erasing gate to determine whether to erase the memory at the $t$-th step as follows:
\begin{equation}
    g^e_t = \sigma(f_{ers}(h_t;\theta_{ers}))
\end{equation}
Considering that the attention weights can indicate the access intensity of each clip feature, we update the video memory with the guidance of that.
Besides the attention weights, to ensure the highly attended features to be erased have actually been described, we further compute their semantic relevance to the generated words as follows:
\begin{eqnarray}
    & \tilde{h}_{t} = \sum_{j=0}^W w_jh_{t-j} \\
    & p^e_{t,i} = \sigma(f_{sem}([\hat m_{t+1,i};\tilde{h}_{t}];\theta_{sem}))
\end{eqnarray}
where $\tilde{h}_{t}$ is the history hidden states computed similar to $\bm{\tilde{\alpha}}_{t}$ in (\ref{eqn:W}).
Finally, the video memories can be updated to $\mathcal{M}_{t+1}$ as follows:
\begin{equation}
    m_{t+1,i} = \hat m_{t+1,i}(1 - g^e_t \tilde{\alpha}_{t,i} p^e_{t,i})
\end{equation}

\subsection{Diversity-driven Training}
\vspace{-2pt}
\label{sec:objective}
The dynamic video memories can help the model to describe diverse video content, while diverse language expression is also essential to the paragraph generation.
The typical MLE and RL training objectives for the captioning model both force the model to fit the ground-truth distribution, which makes the decoder prefer high-frequency tokens and phrases.
It not only results in dull and repetitive expressions, but also makes the model generate wrong descriptions, regardless of video content.
Motivated by the unlikelihood training \cite{welleck2020unlikelihood}, we improve the training objectives with token- and phrase-level high-frequency penalties.

\textbf{Token-level Training.}
In the token-level training, we augment the MLE objective with high-frequency word penalties as \cite{welleck2020unlikelihood}.
Considering that the model tends to repeat words that have been generated before, we define the previous context words as the high-frequency tokens for the current training pair $(v, y^{*})$.
Therefore, the MLE loss function~(\ref{eqn:xe}) is changed to:
\begin{equation}
    \mathcal{L}_{\text{mle}} = -\frac{1}{T}\sum_{t=1}^T(\log p(y^*_t|y^*_{<t},v)+\sum_{c\in C^t}\log(1-p(c|y^*_{<t},v)))
\end{equation}
where $C^t=\{y^*_1,\cdots,y^*_{t-1}\} \setminus \{y^*_t\}$ is the candidate word set to be penalized.
In this way, not only the probabilities of ground-truth words are enhanced, but also the probabilities of wrong candidates with high-frequency are penalized.

\textbf{Sequence-level Training.}
In the sequence-level training, we introduce the phrase-level penalty to the RL loss function~(\ref{eqn:rl}).
Specifically, we compute the $n$-gram frequency of the training annotations to create a phrase frequency look-up table.
The inverse document frequency (idf) score can represent the uniqueness of $n$-grams, which is employed as the diversity reward in reinforcement learning.
To avoid the model generating meaningless phrases which of course have low frequencies in the annotations, we combine the diversity reward with the relevance reward computed by CIDEr \cite{vedantam2015cider}.
The RL loss function~(\ref{eqn:rl}) is changed to:
\begin{equation}
\label{eqn:reward}
    \mathcal{L}_{\text{rl}} = -\frac{1}{T}\sum_{t=1}^T(r_{rlv}(y^s)+\beta r_{div}(y^s_t))\log p(y^s_t|y^s_{<t},v)
\end{equation}
\begin{equation}
\label{eqn:phrase}
    r_{div}(y^s_t) = \frac{1}{k}\sum_{ph \in H_n(y^s_t)}\frac{1}{\mathrm{freq}(ph)}
\end{equation}
where $H_n(y^s_t)$ is the set of $n$-grams containing $y^s_t$ in $y^s$, and $k$ is the size of $H_n(y^s_t)$. $\beta$ is the hyper-parameter to balance the diversity and accuracy.
In practice, we normalize both $r_{rlv}(\cdot)$ and $r_{div}(\cdot)$ with baseline rewards of sentences sampled by greedy search as in \cite{rennie:sc} for training stability.

We first train the whole model with $\mathcal{L}_{\text{mle}}$ and video summary losses as follows:
\begin{equation}
\label{eqn:total_loss}
    \mathcal{L} = \mathcal{L}_{\text{mle}}+\lambda_1\mathcal{L}_{\text{reconst}}+\lambda_2\mathcal{L}_{\text{sparsity}}
\end{equation}
where $\lambda_1$ and $\lambda_2$ are hyper-parameters.
Then we fine-tune the model with $\mathcal{L}_{\text{rl}}$ in reinforcement learning.
\section{Experiments}
\begin{table*}[ht]
\centering
\caption{Comparison with state-of-the-art approaches for video paragraph generation on ActivityNet Captions \emph{ae-test} split. ``Train'' and ``Infer'' indicate if the video segment annotations are needed at training and inference time.}
\label{tab:sota_comparison_1}
\small
\begin{tabular}{l | l |c c| c c c| c c c}
\toprule
 & & \multicolumn{2}{|c}{Segment Annotation} & \multicolumn{3}{|c}{Accuracy} & \multicolumn{3}{|c}{Diversity} \\
\# & Methods & Train & Infer & BLEU@4 & METEOR & CIDEr & Div@1$\uparrow$ & Div@2$\uparrow$ & Rep@4$\downarrow$ \\ 
\midrule
1 & MFT \cite{xiong2018MFT} & \checkmark & \checkmark & 10.33 & 15.09 & 19.56 & - & - & 15.88 \\
2 & VTransformer\footnotemark[3] \cite{zhou2018endtoend} & \checkmark & \checkmark & 10.38 & 16.33 & 21.05 & 61.45 & 77.36 & 7.42 \\
3 & AdvInf\footnotemark[3] \cite{park2019AdvInf} & \checkmark & \checkmark & 10.89 & \textbf{17.41} & 20.40 & 60.59 & 78.29 & 5.09 \\
4 & MART\footnotemark[3] \cite{lei2020MART} & \checkmark & \checkmark & 10.54 & 17.12 & 24.14 & 61.41 & 77.43 & 5.32 \\
\midrule
5 & MFT \cite{xiong2018MFT} & \checkmark & $\bm{\times}$ & 8.45 & 14.75 & 14.15 & - & - & 17.59 \\
\midrule
6 & Vanilla & $\bm{\times}$ & $\bm{\times}$ & 11.53 & 15.91 & 24.11 & 64.92 & 82.34 & 3.17 \\
7 & \textbf{Ours} & $\bm{\times}$ & $\bm{\times}$ & \textbf{12.20} & 16.10 & \textbf{27.36} & \textbf{68.33} & \textbf{84.26} & \textbf{2.63} \\
\midrule
8 & Human & - & - & - & - & - & 68.60 & 85.40 & 0.83 \\
\bottomrule
\end{tabular}
\vspace{-8pt}
\end{table*}

\begin{table}[ht]
\centering
\caption{Captioning results on Charades Captions dataset.}
\label{tab:sota_comparison_2}
\small
\begin{tabular}{l |c c c| c c c}
\toprule
& \multicolumn{3}{|c}{Accuracy} & \multicolumn{3}{|c}{Diversity} \\
Methods & B@4 & M & C & D@1 & D@2 & R@4 \\ 
\midrule
HRL \cite{wang2018hrl} & 18.80 & 19.50 & 23.20 & - & - & - \\
\midrule
Vanilla & 19.19 & 19.80 & 25.30 & 72.90 & 86.13 & 1.23 \\
\textbf{Ours} & \textbf{20.34} & \textbf{20.05} & \textbf{27.54} & \textbf{76.18} & \textbf{87.31} & \textbf{0.92} \\
\midrule
Human & - & - & - &  79.90 & 90.81 & 0.10 \\
\bottomrule
\end{tabular}
\vspace{-10pt}
\end{table}

\subsection{Datasets and Experimental Settings}
\noindent\textbf{Datasets.}
We conduct experiments on the ActivityNet Captions dataset \cite{krishna2017densecap} and Charades Captions dataset \cite{wang2018hrl}.
ActivityNet Captions dataset contains 10,009 videos for training, 4,917 for validation and 5,044 for testing.
Each video in the training set has a single reference paragraph while each video in the validation set has two reference paragraphs.
Since the test set is held for the challenge evaluation, we follow previous works \cite{lei2020MART,zhou2019grounded} to split the validation set into two subsets: \emph{ae-val} with 2,460 videos for validation and \emph{ae-test} with 2,457 videos for test.
Charades Captions dataset is processed from the Charades dataset \cite{sigurdsson2016charades}, which contains 6,963 videos for training, 500 for validation and 1,760 for testing.
Each video is annotated with multiple paragraphs.

\noindent\textbf{Evaluation Metrics.}
We evaluate the paragraph generation qualities from two aspects, \emph{accuracy} and \emph{diversity} respectively.
For the accuracy measurement, we evaluate the generated paragraph against the ground-truth with three standard metrics as \cite{lei2020MART,park2019AdvInf,xiong2018MFT}, including BLEU@4 \cite{papineni2002bleu}, METEOR \cite{denkowski2014meteor} and CIDEr \cite{vedantam2015cider}.
Since the standard metrics do not consider much about diversity of the paragraph, we further evaluate the generated paragraphs with diversity metrics.
Following \cite{park2019AdvInf}, we evaluate the diversity using two types of metrics: 1) $n$-gram diversity (Div@$n$) \cite{shetty2017div1}: the ratio of unique $n$-grams to the total number of words in the paragraph, which is widely used for diversity evaluation; and 2) $n$-gram repetition (Rep@$n$) \cite{xiong2018MFT}: the ratio of $n$-gram repetitions to the total number of $n$-grams.

\noindent\textbf{Implementation Details.}
For the videos, we use ResNet-200 \cite{he2016resnet} pretrained on ImageNet and I3D (RGB+Flow) \cite{carreira2017i3d} pretrained on Kinetics dataset to extract clip-level features of dimensionality 4096D.
We truncate video clips with maximum number of 150.
For the texts in ActivityNet dataset, we truncate the paragraph with maximum length of 150 and build the vocabulary with 10,246 words.
For the Charades dataset, we truncate the paragraph with maximum length of 100 and build the vocabulary with 2,692 words.
We set the number of encoder and decoder layers as $N=3$, the hidden size as $d=512$ and the number of attention heads as 8.
The start window length $S$ in Eq.(\ref{eqn:S}) is set as 50, according to dataset statistics that the first 1/3 length of the video are likely to belong to the first event in the paragraph description.
We set the history window length $W=20$ in Eq.(\ref{eqn:W}) based on the average length of a single sentence.
For the phrase penalty in Eq.(\ref{eqn:phrase}), we set the $n$ to 4.
The $\beta$ in Eq.(\ref{eqn:reward}) is set to 0.3, and the weights of video summary losses in Eq.(\ref{eqn:total_loss}) are set as $\lambda_1=\lambda_2=0.5$.
During training, we use the label smoothing \cite{szegedy2016lablesmooth} with value set as 0.1 and optimize with the learning rate varied under a warm-up strategy with 8,000 steps.
In the inference phase, we generate the paragraph with greedy search.

\begin{table*}[ht]
\centering
\caption{Ablation study on ActivityNet \emph{ae-test} set to demonstrate the effectiveness of different components. (pme: progressive memory exposure, omd: over-accessed memory decay, token: token penalty objective, $r_{rlv}$: relevance reward, $r_{div}$: phrase penalty objective.)}
\label{tab:ablation}
\small
\begin{tabular}{c| c c| c |c c| c c c| c c c}
\toprule
\# & \multicolumn{2}{c|}{Decoder} & MLE & \multicolumn{2}{c|}{RL} & \multicolumn{3}{c|}{Accuracy} & \multicolumn{3}{c}{Diversity} \\ 
& pme & omd & token & $r_{rlv}$ & $r_{div}$ & BLEU@4 & METEOR & CIDEr & Div@1$\uparrow$ & Div@2$\uparrow$ & Rep@4$\downarrow$ \\
\midrule
1 & & & & & & 11.53 & 15.91 & 24.11 & 64.92 & 82.34 & 3.17 \\
\specialrule{0.04em}{0.7pt}{1pt}
2 & \checkmark & & & & & 11.95 & 15.94 & 25.52 & 66.79 & 82.81 & 3.39 \\
3 & & \checkmark & & & & 11.91 & 16.01 & 24.47 & 66.18 & 82.95 & 2.87 \\
4 & \checkmark & \checkmark & & & & 11.61 & 15.72 & 25.65 & 67.90 & 83.37 & 2.80 \\
\specialrule{0.04em}{0.7pt}{1pt}
5 & \checkmark & \checkmark & \checkmark & & & 11.74 & 15.64 & 26.55 & \textbf{68.42} & 83.95 & 2.75 \\
6 & \checkmark & \checkmark & \checkmark & \checkmark & & 12.10 & 15.85 & 27.06 & 67.81 & 83.45 & 2.97 \\
7 & \checkmark & \checkmark & \checkmark & \checkmark & \checkmark & \textbf{12.20} & \textbf{16.10} & \textbf{27.36} & 68.33 & \textbf{84.26} & \textbf{2.63} \\
\bottomrule
\end{tabular}
\vspace{-3pt}
\end{table*}

\begin{table*}[ht]
\centering
\caption{Ablation study on ActivityNet \emph{ae-test} set to demonstrate contributions of different losses for keyframe selection with ratio $\delta=0.5$.}
\label{tab:ablation_loss}
\small
\begin{tabular}{c| c c c| c c c c c c}
\toprule
\# & $\mathcal{L}_\text{mle}$ & $\mathcal{L}_\text{reconst}$ & $\mathcal{L}_\text{sparsity}$ & BLEU@4 & METEOR & CIDEr & Div@1$\uparrow$ & Div@2$\uparrow$ & Rep@4$\downarrow$ \\ 
\midrule
1 & \checkmark & & & 11.60 & 15.51 & 25.47 & 66.97 & 81.06 & 4.30 \\
2 & \checkmark & \checkmark & & 11.41 & 15.58 & \textbf{26.80} & 68.16 & 82.75 & 2.82 \\
3 & \checkmark & & \checkmark & 11.34 & 15.27 & 26.02 & 68.06 & 81.45 & 3.77 \\
4 & \checkmark & \checkmark & \checkmark & \textbf{11.67} & \textbf{15.71} & 26.74 & \textbf{68.71} & \textbf{83.23} & \textbf{2.52} \\
\bottomrule
\end{tabular}
\vspace{-7pt}
\end{table*}

\footnotetext[3]{These strong baselines are rerun using their released codes with the same video features as in our model and are better than the reported results.}

\subsection{Comparison with the State-of-the-arts}
We compare our model with the following state-of-the-art methods, which all use event segments (either ground-truth events or automatically generated events) for video paragraph generation.
\parskip=0.1em
\begin{itemize}[itemsep=1pt,partopsep=0pt,parsep=\parskip,topsep=3pt]
    \item \textbf{MFT} \cite{xiong2018MFT}: A LSTM based model which couples two RNNs for the event detection and event captioning respectively. The previously detected events and generated captions are exploited as context information.
    \item \textbf{VTransformer} \cite{zhou2018endtoend}: A transformer based model which independently generates descriptions for each event segment. We use ground-truth event segments for a stronger baseline as in \cite{lei2020MART}.
    \item \textbf{AdvInf} \cite{park2019AdvInf}: A LSTM based model with hybrid discriminators to select diverse and fluent captions from a sampling set at inference time. The ground-truth event segments are used to generate captions.
    \item \textbf{MART} \cite{lei2020MART}: A transformer based model with memory augmented to fully exploit the event and sentence histories for better captioning. Similar to AdvInf \cite{park2019AdvInf}, the ground-truth event segments are used in generation.
\end{itemize}

Table~\ref{tab:sota_comparison_1} reports the paragraph generation performances of different models on the ActivityNet Captions \emph{ae-test} split.
The table shows that there is a large performance gap for two-stage approaches between using ground-truth event segments (Row 1-4) and using automatically generated ones (Row 5), which demonstrates that the poor performance of event detection seriously hinders the quality of paragraph generation. 
With the advantages of one-stage framework, vanilla model and our final model both outperform these strong baselines and do not require any event segment annotations.
Our model achieves the best results on both the accuracy and diversity aspects except on METEOR.
The METEOR metric prefers longer paragraphs, therefore, the methods using ground-truth event proposals have higher METEOR scores due to the same number of sentences with the ground-truth.
For the diversity aspect, we achieve competitive results with the human level on Div@1 and Div@2, which demonstrates the effectiveness of our proposed dynamic video memories and diversity-driven training objectives.
Experimental results on Charades Captions dataset in Table~\ref{tab:sota_comparison_2} also demonstrate our model achieves state-of-the-art results for video paragraph generation.
Besides automatic metrics, we also conduct human evaluation in the supplementary material to further show the improvements.

\subsection{Trade-off of Efficiency and Performance}
\begin{figure}
  \centering
  \includegraphics[width=0.99\linewidth]{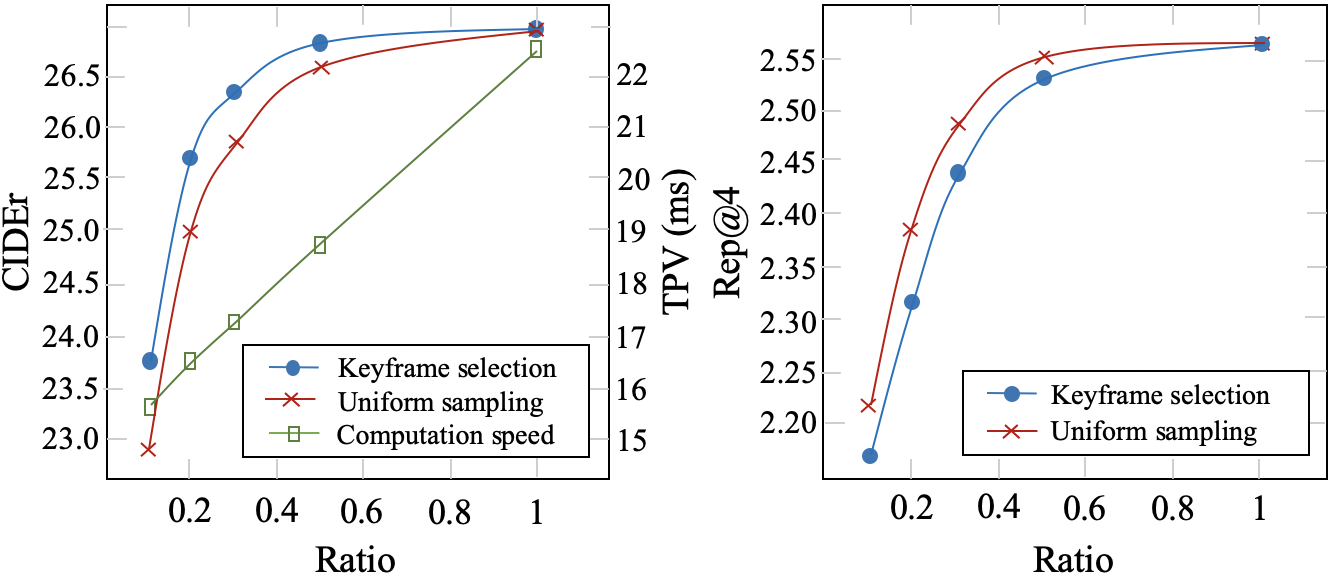}
  \caption{Variation of captioning performance and speed with different selecting ratios $\delta$. TPV denotes time per video (ms).}
  \vspace{-10pt}
  \label{fig:efficiency}
\end{figure}

Figure~\ref{fig:efficiency} shows the inference speed and captioning performances of our model using different keyframe selecting ratios.
The inference speed grows (the time cost drops) rapidly with fewer video clips to be attended to, while the CIDEr drops slowly until the selecting ratio $\delta$ is below 0.4.
It demonstrates that the proposed keyframe selection mechanism is effective which can discard less informative clips in the untrimmed video to improve efficiency and maintain the captioning quality.
The best selecting ratio to trade off the speed and performance is 0.5.
Contrary to the accuracy metric, the diversity of paragraph is improved with fewer frames selected because the video features are more distinctive without redundancy.

We also compare our proposed keyframe selection method with a uniform interval sampling approach, which uniformly selects video clip features instead of relying on the learned $\bm{s}^N$.
Our proposed keyframe selection method outperforms the uniform interval sampling on both CIDEr and Rep@4 under all the selecting ratios, which demonstrates our model can select more distinctive frames.

\begin{figure}[t]
  \centering
  \includegraphics[width=0.98\linewidth]{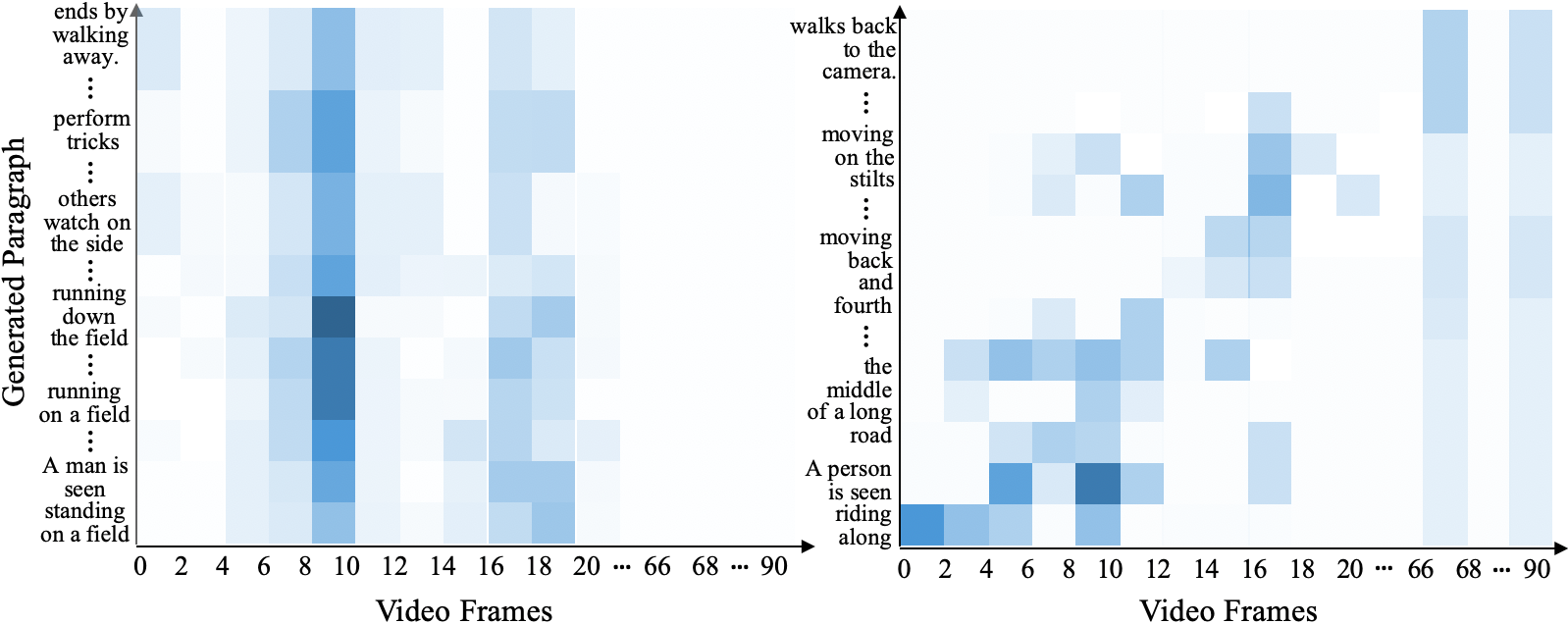}
  \caption{Visualization of the attention weights $\bm{\alpha}$ on video frames for paragraph generation. \emph{Left}: vanilla model. \emph{Right}: our model with dynamic video memories.}
  \vspace{-18pt}
  \label{fig:attention}
\end{figure}

\begin{figure*}
    \centering
    \begin{overpic}[scale=0.49]{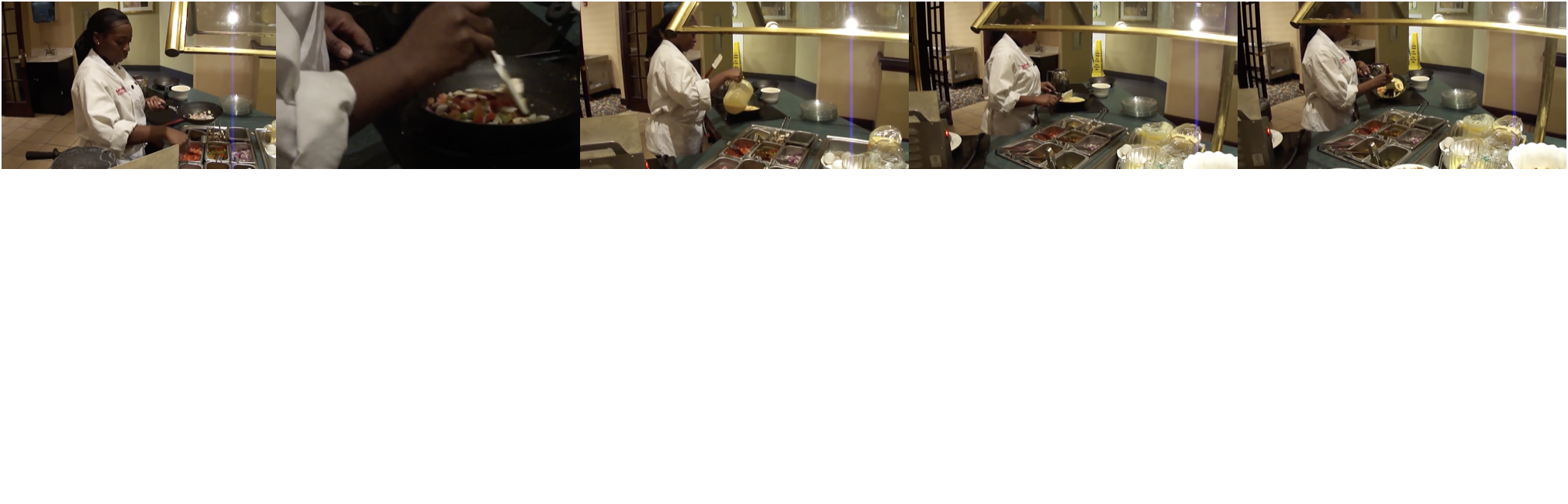}
        \put(0.2,19.4){\footnotesize{\textbf{\underline{VTransformer} (GT events)}: He starts cooking in the kitchen. A chef is standing at a counter in a kitchen. {\color[RGB]{220,20,60}A man} is standing in a kitchen.}}
        \put(0.2,17.3){\footnotesize{\textbf{\underline{AdvInf} (GT events)}: {\color[RGB]{220,20,60}A man} is standing in front of a counter while {\color[RGB]{220,20,60}speaking to the camera} and leads into him taking a pan and presenting it to the camera. }}
        \put(0.2,15.4){\footnotesize{The camera pans around the food and ends by presenting it to the camera. {\color[RGB]{220,20,60}The man} takes a sip of the food and begins to stir the pan.}}
        \put(0.2,13.4){\footnotesize{\textbf{\underline{MART} (GT events)}: {\color[RGB]{220,20,60}A man} is seen standing behind a table {\color[RGB]{220,20,60}speaking to the camera} and begins mixing ingredients into a pan. {\color[RGB]{220,20,60}The man} continues to mix }}
        \put(0.2,11.6){\footnotesize{ingredients and ends by presenting it to the camera. He continues {\color[RGB]{220,20,60}speaking to the camera} and showing off his finished in the end.}}
        \put(0.2,9.6){\footnotesize{\textbf{\underline{Vanilla} (no event detection)}: {\color[RGB]{220,20,60}A man} is seen standing behind a counter {\color[RGB]{220,20,60}speaking to the camera} and leads into him holding up a food. {\color[RGB]{220,20,60}The man} then mixes }}
        \put(0.2,7.8){\footnotesize{ingredients into a bowl and finally putting food into a pan.}}
        \put(0.2,5.8){\footnotesize{\textbf{\underline{Ours} (no event detection)}: A woman is seen standing behind a counter and putting various ingredients into a pan. She mixes the ingredients together and }}
        \put(0.2,4){\footnotesize{ends by spreading it onto a plate.}}
        \put(0.2,2){\footnotesize{\textbf{\underline{Ground-Truth}}: A woman is seen cooking items onto a stove with various ingredients laid out. The camera pans around kitchen and shows the woman }}
        \put(0.2,0.2){\footnotesize{cooking more ingredients. She continues mixing it around in the pan.}}
    \end{overpic}
    \caption{Qualitative examples of the generated paragraphs by our model and other state-of-the-arts methods. The words in {\color[RGB]{220,20,60}red} represent high-frequency tokens and phrases which are generated regardless of video content.}
    \vspace{-13pt}
    \label{fig:results}
\end{figure*}

\subsection{Ablation Studies}
In order to demonstrate the contributions from different components in our model, we carry out ablation studies in Table~\ref{tab:ablation}.
Row 1 denotes the vanilla baseline presented in Section~\ref{sec:background}, which directly generates the video paragraph as in conventional video captioning task.
In Row 2-4, we replace the standard attention mechanism in the vanilla transformer decoder with our proposed attention enhanced by dynamic video memories.
The proposed progressive memory exposure and over-accessed memory decay mechanisms both improve the paragraph diversity by 1-2 points on Div@1,
while combining them achieves additional gains on both accuracy and diversity metrics. 
We visualize the attention weights from vanilla model and our proposed model in Figure~\ref{fig:attention}. It shows the learned attentions in vanilla model only focus on a few salient clips for the whole paragraph, which leads to repeated or missed event descriptions. However, with our proposed dynamic video memories according to the description status, our attention can focus on diverse frames and roughly forms a diagonal line similar to human description with the chronological order.
In Row 5, we add the token-level high frequency penalty to the MLE training objective, which further brings improvements.
In Row 6 and Row 7, we finetune the pretrained model via reinforcement learning.
With the CIDEr reward alone, the model yields better accuracy results, but harms diversity metrics.
Combining it with phrase-level high frequency penalty achieves the best final results.
The proposed diversity-driven training objectives (both token- and phrase-level) are shown not only improving the diversity metrics but also the accuracy metrics due to their good abilities to prevent the language decoder from generating high-frequency words regardless of visual content.

Table~\ref{tab:ablation_loss} shows the effectiveness of the two auxiliary losses in the keyframe selection.
We report the results with 50\% video clips selected in inference since it is the best ratio choice to trade off the speed and performance.
Without the video reconstruction loss, the model learns to select keyframes only with the captioning loss, which results in a large performance drop.
However, the proposed reconstruction loss can help to enforce the model to select salient frames that maintain similar semantic information with the original video clip features.
Our final loss with the three losses combined achieves the best results.

\subsection{Qualitative Analysis}
Figure~\ref{fig:results} shows a test example with the paragraph captions from our model and other state-of-the-art models.
The compared models incorrectly describe the woman as ``man'' due to its higher frequency (man 2.26\% vs. woman 0.85\%) in the data.
Furthermore, they tend to generate redundant high-frequency phrases regardless of the video content, such as ``speaking to the camera'', which is the top1 frequent verb phrase in the training set.
Our model, however, can generate more coherent and diverse video paragraphs even without ground-truth event segment annotations.
More examples can be found in the supplementary material.
\vspace{-2pt}
\section{Related Works}
Over the past years, image captioning has achieved significant improvements \cite{vinyals2015showtell,xu2015showatttell,you2016semanticatt,lu2017adaptiveatt,anderson2018butd}, which mainly focus on generating a single sentence to describe the image content.
In order to describe more fine-grained details in an image, Krause \etal \cite{krause2017imageparagraph} propose the image paragraph captioning with a hierarchical RNN to generate topic vectors first and then convert topics into sentences to form a paragraph.
However, recent works \cite{luo2019paragraph,Melas2018paragraph} have shown that directly generating paragraph as a long sentence outperforms the hierarchical manner when enhanced with diversity-driven training and inference approaches.
Inspired by the image paragraph generation pioneers, in this work we explore whether the video paragraph can be effectively generated without hierarchical manner of using event detection.

Video captioning \cite{venugopalan2015vtt,yu2017endtoend,wang2018m3,chen2019topic,wang2018reconstruction} is more challenging compared to image captioning with complexities on both spatial and temporal dimensions.
Recently, Krishna \etal \cite{krishna2017densecap} propose the dense video captioning task to localize and describe multiple events for long videos with multiple sentences.
They first detect multiple events in the video and then generate description for each of them.
However, these descriptions are independent and not coherent as a whole.
Xiong \etal \cite{xiong2018MFT} further propose to generate a coherent paragraph to describe multiple activities in the long video.
However, they still solve the problem in a two-stage way like dense video captioning methods \cite{zhou2018endtoend,wang2018bidirectional,li2018jointly,mun2019streamline}.
They first propose hundreds of event proposals with event proposal networks \cite{zhao2017TAD,buch2017sst,escorcia2016DAPs}, then select events to be described with the contextual information from previously detected events and generated captions.
%However, there is still a large performance gap between generating with the learned events and that with the ground-truth events.
Park \etal \cite{park2019AdvInf} and Lei \etal \cite{lei2020MART} directly generate video paragraphs with the ground-truth event segments, which cannot generalize to videos without such event annotations.
In this work, we eschew the event detection stage and directly generate the video paragraph with dynamic video memories.
\vspace{-2pt}
\section{Conclusion}
In this paper, we propose an one-stage framework for video paragraph generation.
Due to the long video inputs and paragraph outputs, it is challenging to generate diverse paragraphs efficiently.
We propose a keyframe-aware video encoder to improve the efficiency and an attention mechanism with dynamic video memories to learn more diverse and coherent visual attentions.
Besides, a diversity-driven training objective with high-frequency token and phrase penalties is proposed to improve language diversity.
Experimental results on ActivityNet and Charades datasets show that our proposed model outperforms the state-of-the-art performance on both accuracy and diversity metrics.

\vspace{-2pt}
\section*{Acknowledgment}
\vspace{-1pt}
This work was supported by National Natural Science Foundation of China (No. 62072462), Beijing Natural Science Foundation (No. 4192028), and National Key R\&D Program of China (No. 2020AAA0108600).

{\small
\bibliographystyle{ieee_fullname}
\bibliography{egbib}
}

\end{document}